\documentclass[conference]{IEEEtran}
\IEEEoverridecommandlockouts
\AtBeginDocument{%
  }

\usepackage{stfloats}
\usepackage{verbatim}

\usepackage{soul}

\usepackage{tikz}
\usepackage{amsmath}

\usepackage{filecontents}
\usepackage{graphicx}
\usepackage{textcomp}
\usepackage{academicons}
\usepackage{xcolor}
\usepackage{float}
\usepackage{algorithm}
\usepackage{algpseudocode}

\usepackage{todonotes}
\usepackage[utf8]{inputenc}

\usepackage[normalem]{ulem}
\usepackage{amsmath}
\usepackage{amsfonts}
\usepackage{tabularray}
\usepackage{lscape}
\usepackage{mathtools}
\usepackage{booktabs}
\usepackage{color}
\usepackage{xcolor}
\usepackage{caption}
\usepackage{subcaption}
\usepackage{algpseudocode}
\usepackage{algorithm}

\usepackage{enumitem}
\usepackage{array}
\usepackage{tikz}
\usepackage{soul}

\newcolumntype{L}[1]{>{\raggedright\let\newline\\\arraybackslash\hspace{0pt}}m{#1}}
\newcolumntype{C}[1]{>{\centering\let\newline\\\arraybackslash\hspace{0pt}}m{#1}}
\newcolumntype{R}[1]{>{\raggedleft\let\newline\\\arraybackslash\hspace{0pt}}m{#1}}

\begin{document}
\pagestyle{plain} 

\title{Faults That Fortify: CNN Adversarial Robustness via GPU Undervolting}

\author{
\IEEEauthorblockN{Behnam Omidi\IEEEauthorrefmark{1},Ahmad Tahmasivand\IEEEauthorrefmark{1},
Husam Alsyouri\IEEEauthorrefmark{1},
Saba Al-Sayouri\IEEEauthorrefmark{2},
Chongzhou Fang\IEEEauthorrefmark{3}}
\IEEEauthorblockN{Ihsen Alouani\IEEEauthorrefmark{4},and
Khaled N. Khasawneh\IEEEauthorrefmark{1}}


\IEEEauthorblockA{\IEEEauthorrefmark{1}George Mason University,
Fairfax, USA - 
Email: \{bomidi, atahmasi, halsyour, kkhasawn\}@gmu.edu}
\IEEEauthorblockA{\IEEEauthorrefmark{2}SecureMind Technologies Inc,
Fairfax, USA - 
Email: saba@securemindtech.ai}
\IEEEauthorblockA{\IEEEauthorrefmark{3}Rochester Institute of Technology,
Rochester, USA - 
Email: cxfeec@rit.edu}


\IEEEauthorblockA{\IEEEauthorrefmark{4}Queen's University, Belfast, UK - 
Email: i.alouani@qub.ac.uk}
}

\maketitle

\begin{abstract}
Convolutional Neural Networks (CNNs) face a dual challenge: vulnerability to adversarial attacks and prohibitive training cost. Adversarial training is effective but expensive, a burden that grows as learning shifts to the energy-constrained edge. This paper addresses both through GPU undervolting during training. Reducing supply voltage introduces stochastic perturbations in arithmetic that act as implicit regularization, improving robustness while lowering power. We characterize undervolting-induced faults at the bit level, then train LeNet, VGG-6, and MobileNetV3 on MNIST and CIFAR-10 under two training regimes, standard and adversarial, each at nominal and undervolted voltage, and evaluate all models against adversarial attacks. In both regimes, the undervolted model consistently achieves higher adversarial accuracy than its nominal-voltage counterpart, showing that hardware-induced faults strengthen even adversarial training. Because dynamic power scales quadratically with supply voltage, these robustness gains arrive with substantial energy savings. GPU undervolting is therefore a readily deployable hardware-level defense requiring no algorithmic change, and opens a promising direction in which robustness and energy efficiency move together. 
\end{abstract}


\maketitle

\section{Introduction}
\label{intro}

Convolutional Neural Networks (CNNs) have become a fundamental component of modern computer vision, enabling applications ranging from medical imaging to autonomous driving. However, training these models requires substantial computational resources, including high-performance GPUs and significant energy consumption \cite{patterson2021carbon}. These challenges are particularly pronounced in edge computing, where models are increasingly fine-tuned in addition to being deployed on resource-constrained platforms such as drones, wearable devices, and autonomous vehicles operating under stringent power budgets \cite{lin2022device}. In such settings, the computational overhead of robustness-enhancing techniques is often prohibitive. Consequently, improving training efficiency while preserving both model accuracy and robustness has emerged as a critical research objective.

\begin{figure}[ht!]
    \centering
    \includegraphics[width=1\linewidth]{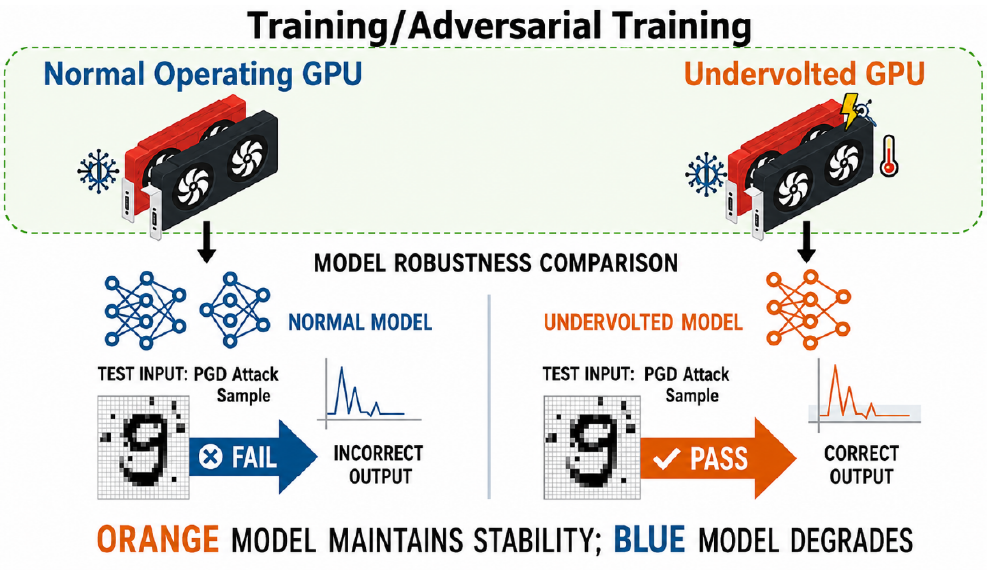}
    \caption{Workflow overview}
    \label{fig:placeholder}
\end{figure}

Recent research focuses on the algorithmic level through quantization, pruning, and accelerator codesign \cite{menghani2023efficient, gholami2022survey}. A complementary lever acts beneath it, where hardware modifications such as voltage scaling influence model performance and learning dynamics \cite{leng2015safe}. Prior works have examined the security implications of this behavior, showing how undervolting can be exploited to induce faults in controlled ways \cite{juffinger2024suit, tang2017clkscrew}, while also demonstrating its potential as a defensive mechanism that injects benign noise to disrupt adversarial attacks \cite{majumdar2021using, islam2023vpp,
islam2021lower}. These undervolting-induced faults result from diminished voltage margins that degrade transistor switching reliability, potentially causing timing violations, unstable storage behavior, and incorrect signal sampling during computation. Unlike memory-centric fault mechanisms, such as Rowhammer-induced DRAM disturbances or targeted weight corruption attacks \cite{coalson2024prisonbreak, zahran2025jailbreaking, abharian2026gbfa}, which involve persistent alterations to stored data and are addressed through corresponding mitigation approaches \cite{tahmasivand2025lm, nazari2024forget}, undervolting faults arise from transient disruptions in the underlying hardware execution process.

It is also worth noting that these defensive mechanisms primarily operate during inference, where faults are introduced into an already trained model and are typically emulated through software-based injection \cite{islam2021lower, majumdar2021using} rather than experimentally observed under undervolted hardware conditions. Consequently, the impact of transient faults during the training phase remains largely unexplored. In contrast, this work investigates transient faults induced exclusively through undervolting a physical GPU and examines whether such perturbations can be leveraged to beneficially influence training dynamics.

Undervolting injects subtle stochasticity into the arithmetic of learning itself, which may push the network toward flatter representations rather than the sharp decision boundaries that gradient-based attacks exploit. Because dynamic power scales quadratically with supply voltage, the same knob also lowers energy consumption. The technique therefore carries a dual benefit, greener training and more resilient models, from one hardware setting requiring no algorithmic change.


To investigate this, we run standard and adversarial training, each at nominal and undervolted GPU voltage, training LeNet~\cite{lecun1998mnist}, VGG-6~\cite{simonyan2015very}, and MobileNetV3~\cite{howard2019searching} on MNIST~\cite{lecun1998mnist} and CIFAR-10~\cite{krizhevsky2009learning} and evaluating all models against PGD attacks~\cite{madry2018towards} across a range of perturbation budgets (Figure~\ref{fig:placeholder}). In both regimes, the undervolted variant is consistently more robust than its nominal-voltage counterpart while consuming less energy.

These findings represent an encouraging first step rather than a complete defense. Beyond demonstrating that undervolting can improve robustness against adversarial attacks while simultaneously offering energy-efficiency benefits, our results suggest a promising new research direction at the intersection of hardware reliability, energy optimization, and AI robustness. Building on these encouraging results, future work in this area is encouraged to further explore bit-level fault behavior and extend evaluations to larger foundation and quantized models, where both robustness dynamics and energy savings may become even more pronounced. The contributions of this paper are summarized as follows.

\begin{itemize}
\item We characterize the computational faults induced by undervolting on a real GPU, showing that they are stochastic yet controllable.

\item We propose training-time undervolting as a novel means of improving CNN robustness on GPUs, demonstrating that hardware-generated noise during training is a practical, easily deployable, and energy-efficient security mechanism.

\item We evaluate undervolting under both standard and adversarial training and show that it improves robustness in both regimes.

\item We quantify the resulting power reduction, showing that these robustness gains arrive alongside measurable energy savings and acceptable computational reliability.
\end{itemize}

\section{Background and Related Work}
\label{background}

\subsection{Undervolting-Induced Faults}

Undervolting, the reduction of a circuit's supply voltage below its nominal
level, is an effective way to improve energy efficiency in high-performance
systems such as GPUs, since dynamic power scales approximately with the square
of the supply voltage. The same reduction, however, slows transistor switching
and degrades circuit timing, producing computational instability and transient
faults. The quantitative relation between voltage, delay, and error probability
therefore determines how undervolting affects GPU-based deep learning.

\noindent
\textbf{Impact on Power Consumption}: Total GPU power is the sum of dynamic and
static components,

\begin{equation}
P_{\text{total}} = P_{\text{dyn}} + P_{\text{static}}
= \alpha_s C_L V_{dd}^{2} f_{\text{clk}} + I_{\text{leak}} V_{dd},
\end{equation}

\noindent
where \( \alpha_s \) is the switching activity factor, \( C_L \) the effective
load capacitance, \( f_{\text{clk}} \) the clock frequency, and
\( I_{\text{leak}} \) the leakage current. Because \( P_{\text{dyn}} \) scales
quadratically with \( V_{dd} \) while \( P_{\text{static}} \) decreases
linearly, lowering the voltage reduces both terms, at the cost of timing
reliability.

\noindent
\textbf{Voltage-Delay Relationship and Timing Faults}: The propagation delay
\( t_d \) of a CMOS logic gate depends strongly on the supply voltage,

\begin{equation}
t_d \propto \frac{V_{dd}}{(V_{dd} - V_{th})^{\alpha}},
\qquad
t_d \leq T_{\text{clk}} = \frac{1}{f_{\text{clk}}},
\end{equation}

\noindent
where \( V_{th} \) is the threshold voltage, \( \alpha \) (typically between 1
and 2) the velocity saturation index, and \( T_{\text{clk}} \) the clock period.
Correct operation requires the second condition to hold on every logic path.
When undervolting pushes \( t_d \) beyond \( T_{\text{clk}} \), \textit{timing
violations} occur and produce incorrect outputs or transient faults. The
likelihood of such faults grows exponentially as the voltage margin shrinks,

\begin{equation}
P_{\text{fault}} \approx e^{-\beta (V_{dd} - V_{\text{min}})},
\end{equation}

\noindent
where \( V_{\text{min}} \) is the minimum stable voltage and \( \beta \) a
hardware-dependent sensitivity constant. As \( V_{dd} \) approaches
\( V_{\text{min}} \), the system transitions sharply from reliable computation
to frequent soft errors. In GPU workloads these faults appear mainly as small
perturbations in floating-point arithmetic,

\begin{equation}
\tilde{y} = y + \delta(V_{dd}),
\end{equation}

\noindent
where \( y \) is the correct output, \( \tilde{y} \) the computed result under
undervolting, and \( \delta(V_{dd}) \) voltage-dependent noise whose variance
grows as the voltage is lowered. Mild undervolting therefore introduces
stochastic noise, whereas aggressive undervolting causes substantial numerical
instability.

\noindent
\textbf{Implications for Deep Learning}: Such faults are not always harmful.
Neural networks tolerate low-level noise because of their distributed
representations, and the random errors induced by undervolting may act as a
regularizer comparable to dropout or weight noise, potentially improving
generalization and robustness. Excessive undervolting, in contrast, disrupts
gradient-based optimization and degrades convergence and accuracy.

\subsection{Randomization-based Defenses}

\noindent
Convolutional Neural Networks (CNNs) remain vulnerable to adversarial
perturbations, small crafted input changes that drastically alter predictions.
Randomization-based defenses counter this threat by injecting
stochasticity into inference or training so that the structured gradients
attackers rely on become unreliable. Randomness has been applied at three
levels: to inputs, through stochastic resizing, padding, and augmentation
~\cite{xie2017mitigating,dhillon2018stochastic}; to intermediate features,
through noise injection that turns a single network into an implicit ensemble
of stochastic models~\cite{liu2018towards}; and to parameters, through Monte
Carlo dropout, Bayesian uncertainty modeling, and stochastic quantization or
weight noise~\cite{hong2022certified}. Randomized
smoothing formalizes the idea by averaging predictions under Gaussian noise to
obtain probabilistic robustness guarantees within a norm-ball
~\cite{lecuyer2019certified,cohen2019certified}, with later variants improving
certified robustness at comparable
accuracy~\cite{hao2022gsmooth,hong2022certified}.

\noindent
The paradigm has known limits. Expectation-over-transformation attacks recover
usable gradients by averaging over the randomization
itself~\cite{athalye2018obfuscated}, and excessive noise degrades clean-input
accuracy, so the degree of stochasticity requires calibration. Randomization nevertheless remains a flexible and inexpensive route to robustness, and hardware-level stochastic effects such as those induced by GPU undervolting extend it by supplying intrinsic randomization during training at no additional computational cost, while leaving inference deterministic.
\section{Proposed Methodology}
\label{proposed_method}

We propose a training methodology that exploits hardware-level
undervolting to inject beneficial stochastic noise into CNN training.
This section defines the threat model, formalizes how
undervolting-induced faults enter the learning process, and describes
the training procedure and operating-point selection.

\subsection{Threat Model}
\label{sec:threat_model}

We consider the standard white-box evasion setting: the adversary has
full knowledge of the trained model's architecture and parameters and
crafts $L_p$-bounded perturbations using first-order methods,
instantiated in our evaluation as PGD over a range of budgets
$\epsilon$. 
The adversary attacks the deployed model, which executes at nominal
voltage; the defender controls only the hardware configuration of the
training process. We assume the defender's GPU supports voltage
adjustment through standard driver tooling and that training completes
without fatal faults, a condition enforced by the operating-point
selection in Section~\ref{sec:procedure}. Fault-injection attacks on
the model itself~\cite{coalson2024prisonbreak, zahran2025jailbreaking,
abharian2026gbfa} are outside our scope.

\subsection{Undervolting as a Training-Time Noise Source}
\label{sec:noise_source}

As established in Section~\ref{background}, operating below
nominal voltage perturbs individual arithmetic operations as
$\tilde{y} = y + \delta(V_{dd})$, where the variance of $\delta$ grows
as voltage decreases.

During training, every operation in the forward pass, backward pass,
and weight update is subject to such perturbation. The parameter
update at step $t$ is therefore effectively
\begin{equation}
    w_{t+1} \;=\; w_t \;-\; \eta\left(\nabla L(w_t) + \delta_t\right),
    \label{eq:noisy_update}
\end{equation}
where $\eta$ is the learning rate, $L$ the training loss, and
$\delta_t$ aggregates the voltage-induced perturbations accumulated
through the computation of the gradient. Unlike software-based noise
injection, which adds computation and targets a chosen site (inputs,
activations, or weights), $\delta_t$ arises in every operation at zero
computational cost and with no modification to the training pipeline.
Rather than suppressing these perturbations as errors, we treat them
as intrinsic noise regularization, consistent with prior evidence that
stochastic noise applied during training improves adversarial
robustness~\cite{lecuyer2019certified}.
Because per-operation perturbations concentrate in low-significance
mantissa bits (Section~\ref{sec:fault_characterization}), they are
small enough for optimization to converge to full clean accuracy, yet
pervasive enough that their cumulative effect shapes the learned
solution. At deployment, the model executes at nominal voltage; no
faults occur at inference and the model's outputs are fully
deterministic.

\subsection{Training Procedure and Operating-Point Selection}
\label{sec:procedure}

For each architecture--dataset pair, we sweep voltage--frequency
operating points downward from nominal and select the lowest voltage
at which training reliably completes, the point of maximum fault
incidence short of instability. The GPU is held at this voltage for
the entire training run, so perturbations act on every iteration from
initialization to convergence.

We evaluate the method under a $2{\times}2$ design: two training
regimes, standard and adversarial, each executed at nominal and
undervolted voltage with matched random seeds, weight
initializations, and hyperparameters, so that within each regime the
applied voltage is the only systematic difference between runs.
Adversarial noise is deliberately crafted to maximize model error
within a constrained threat model, whereas undervolting noise arises
from hardware fluctuations outside the nominal voltage margin;
because the two differ in both origin and statistical properties, we
hypothesize that they act as complementary robustness mechanisms. In
the adversarial regime, on-the-fly PGD example generation executes
under the same voltage condition as the weight updates, since both
are components of the training process. All trained models are
evaluated identically at nominal voltage.

\section{Experimental Setup}
\label{setup}

All experiments used \texttt{PyTorch} on a dedicated server with an NVIDIA
\texttt{RTX 3090} GPU, which provides 10,496 CUDA cores and 24~GB of GDDR6X
memory. Undervolting was applied through \texttt{MSI Afterburner}, which
exposes voltage in 6~mV steps and frequency in 15~MHz steps, fine enough to
explore the energy-accuracy trade-off across operating points systematically.

We evaluated three architectures spanning a range of depths and computational
budgets: the classical \texttt{LeNet}, a lightweight \texttt{VGG-6}, and
\texttt{MobileNetV3}. Training used two benchmark image-classification datasets,
\textit{MNIST} and \textit{CIFAR-10}. MNIST offers a clean, low-complexity
setting for controlled observation of basic behavior, while CIFAR-10 demands
deeper feature extraction and exercises undervolting under heavier
computational load. To evaluate the robustness of the model, we employed the untargeted PGD attack implemented in the \texttt{torchattacks} version $3.5.1$ library with $\alpha = 0.007$ and $100$ optimization steps, varying the perturbation budget $\epsilon$ until the model accuracy approached zero.
\section{Evaluation}
\label{eval}

\subsection{Characterizing Undervolting-Induced Faults}\label{sec:fault_characterization}

Before evaluating training outcomes, we characterize the faults undervolting induces on our GPU, since these faults are the noise source our method relies on. Prior characterizations were performed on CPUs~\cite{islam2021lower} and FPGAs~\cite{salami2020experimental, islam2023vpp}, where undervolting measurably perturbs inference computation; whether GPU faults behave comparably has not been examined.

\begin{figure*}[!ht]
\centering
        \captionsetup[subfigure]{font=Huge}
	    \resizebox{\textwidth}{!}{

            \begin{subfigure}[b]{\textwidth}
            \centering
            \includegraphics[width=\textwidth]{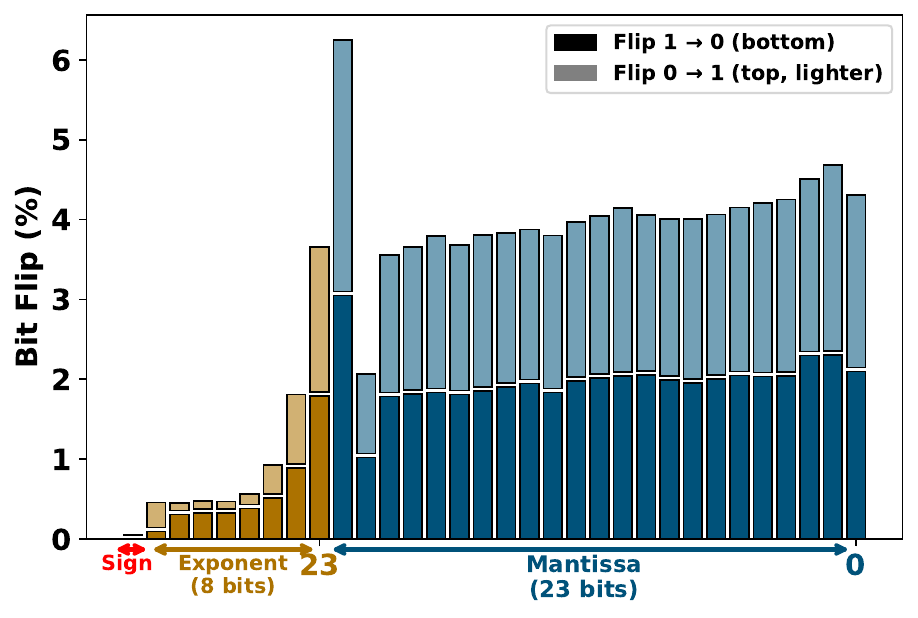}
            \caption{Bit flip distribution across bit position}
            \label{fig:bitflip_distribution} 
           \end{subfigure}

           \begin{subfigure}[b]{\textwidth}
              \centering
              \includegraphics[width=\textwidth]{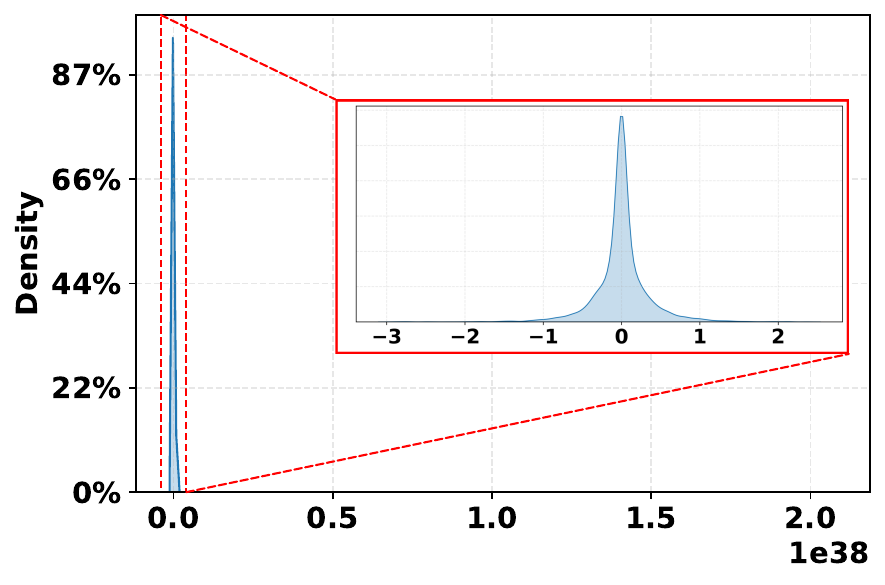}
              \caption{Diversity}
              \label{fig:density}
           \end{subfigure}

            \begin{subfigure}[b]{\textwidth}
              \centering
              \includegraphics[width=\textwidth]{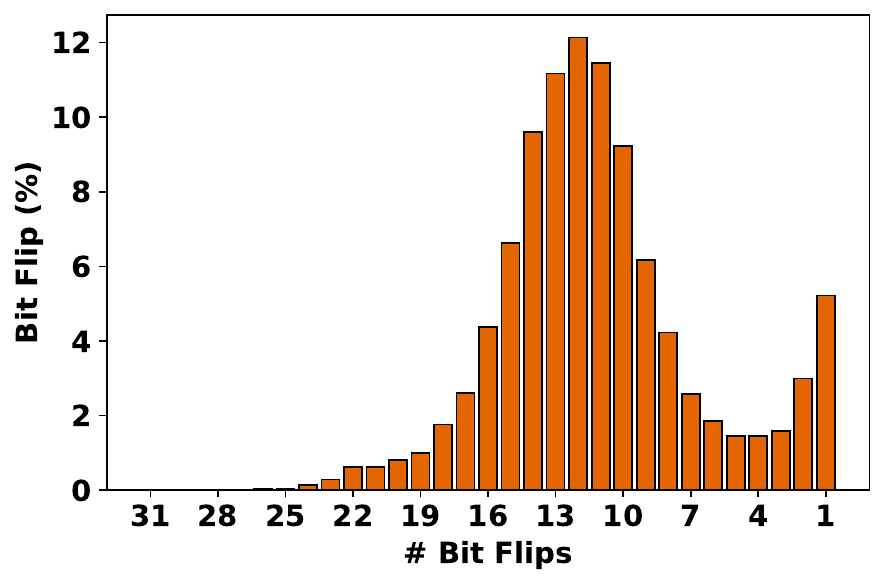}
              \caption{Number of bit flips}
              \label{fig:number_of_bitflip_distribution} 
           \end{subfigure}}
           
    \caption{Impact of undervolting on multiplication in GPU}
    \label{fig:undervolting_mul_gpu}
\end{figure*}

We recreated the core experiment of~\cite{islam2021lower}, originally conducted on a CPU, multiplying two tensors repeatedly, a billion times in our case, to amplify faults until bit-level errors become measurable. We ran the computation at nominal voltage and at the lowest stable undervolted setting the hardware supports, then compared outputs bit by bit, tracking the number of flips and their distribution across the 32-bit floating-point representation. Figure~\ref{fig:undervolting_mul_gpu} reports the frequency of flips and the direction of each transition per bit position.

\begin{figure*}[!ht]
\centering
        \captionsetup[subfigure]{font=Huge}
	\resizebox{\textwidth}{!}{

            \begin{subfigure}[b]{\textwidth}
            \centering
            \includegraphics[width=\textwidth]{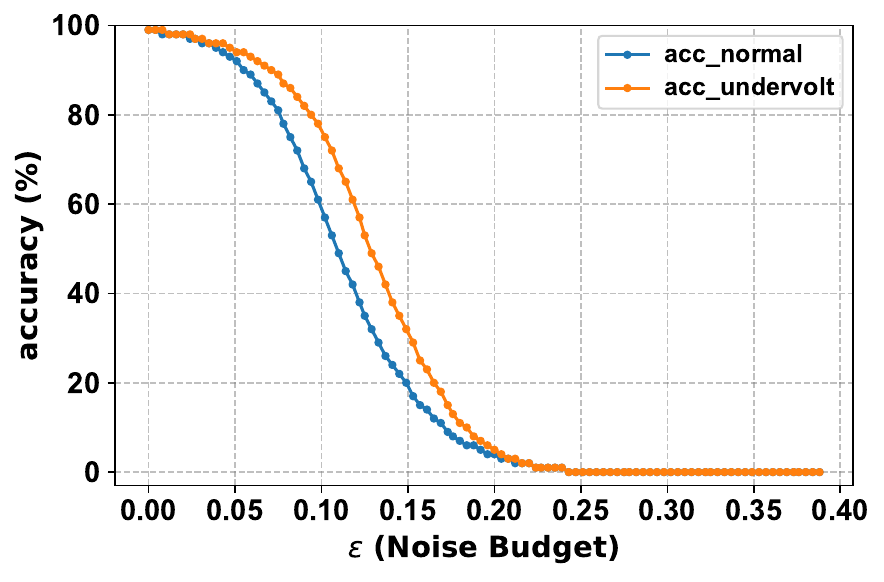}
            \caption{LeNet-MNIST}
            \label{fig:lenet_mnist} 
           \end{subfigure}
           
            \begin{subfigure}[b]{\textwidth}
              \centering
              \includegraphics[width=\textwidth]{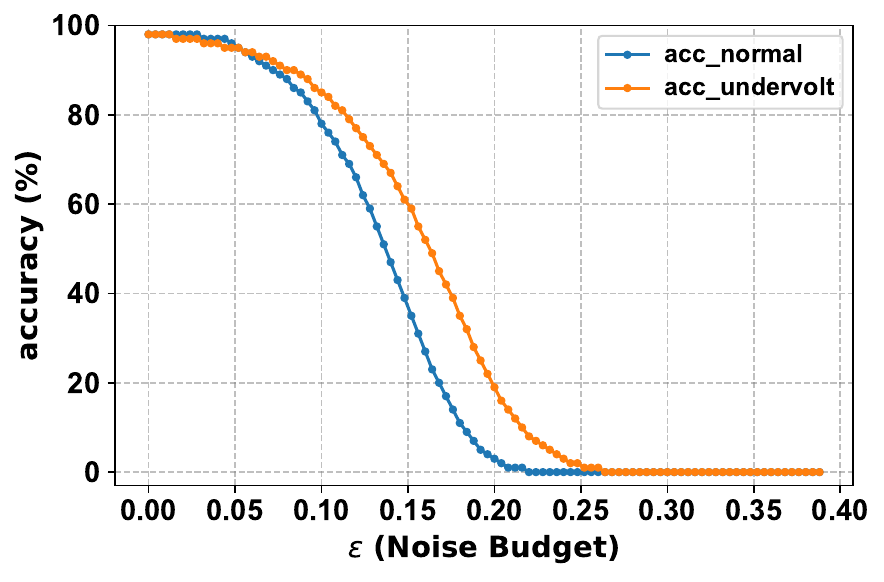}
              \caption{VGG-6-MNIST}
              \label{fig:vgg6_mnist} 
           \end{subfigure}
           
           \begin{subfigure}[b]{\textwidth}
                \centering
                \includegraphics[width=\textwidth]{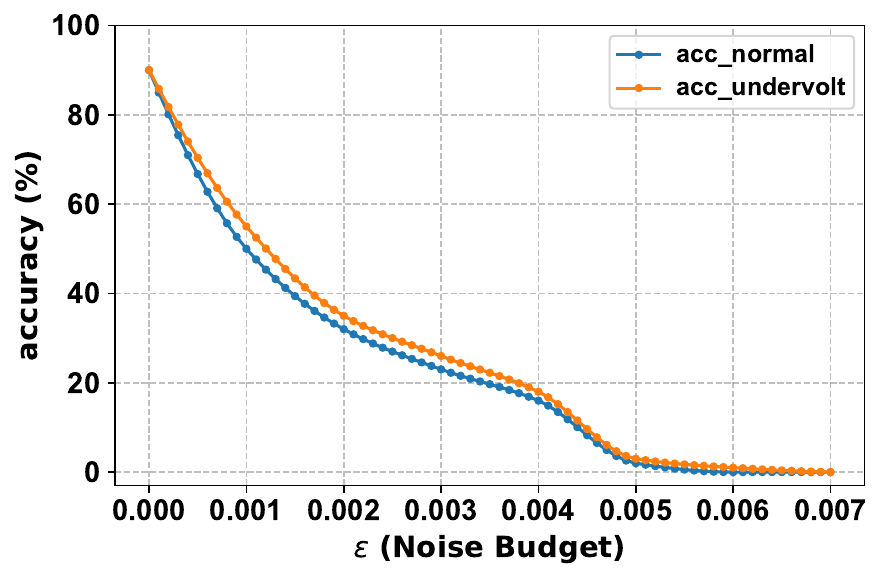}
                \caption{MobileNet-V3-CIFAR-10}
                \label{fig:bitflip_distribution_1_0} 
           \end{subfigure}}
           
    \caption{Comparison of the robustness of original models trained under normal GPU operation and undervolted GPU conditions using adversarial samples generated with various noise budgets}
    \label{fig:undervolting_training}
\end{figure*}

Nearly all observed flips fall in the mantissa, with far fewer in the exponent field. Mantissa bits encode fine-grained precision rather than large numeric shifts, so each fault perturbs a result by an amount well below the sensitivity threshold of a trained network. Consistent with this, we verified that a trained model's inference accuracy is nearly identical at nominal and undervolted voltage on our GPU, in contrast to the degradation reported on other platforms~\cite{salami2020experimental}; individual faults are simply too small to alter predictions in a single forward pass.

This characterization explains why undervolted training remains viable and beneficial. Because per-operation perturbations are small, stochastic, and unbiased, gradient descent tolerates them and converges to full clean accuracy. Because they are injected into every arithmetic operation across millions of training iterations, their cumulative effect nonetheless shapes the optimization, acting as the implicit regularization whose impact on robustness we evaluate next.

\subsection{Undervolted Training}

Figure~\ref{fig:undervolting_training} contrasts standard and undervolted training
under adversarial perturbation. For each model we first trained conventionally
at nominal GPU voltage, then generated adversarial samples with PGD, sweeping over a range of
perturbation budgets. The resulting accuracies
form the blue curve in the figure and show the expected degradation as
perturbation strength grows.

\begin{figure*}[!ht]
\centering
        \captionsetup[subfigure]{font=Huge}
	\resizebox{\textwidth}{!}{
            \begin{subfigure}[b]{\textwidth}
            \centering
            \includegraphics[width=\textwidth]{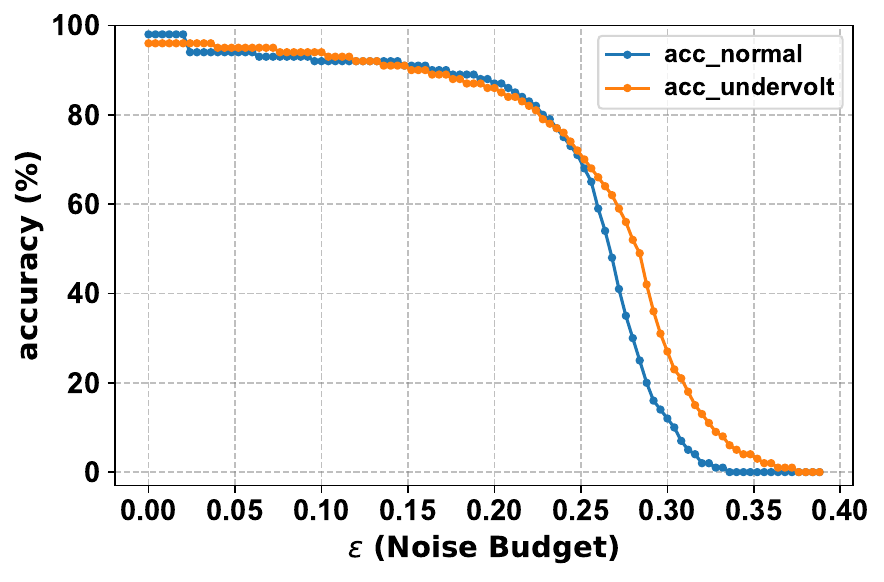}
            \caption{LeNet-MNIST}
            \label{fig:at_lenet_mnist} 
           \end{subfigure}
           
            \begin{subfigure}[b]{\textwidth}
              \centering
              \includegraphics[width=\textwidth]{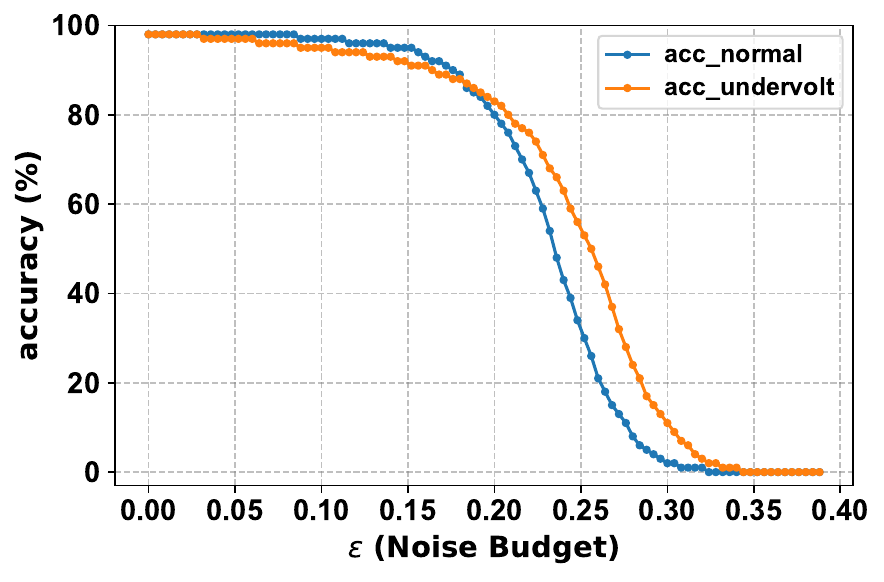}
              \caption{VGG-6-MNIST}
              \label{fig:at_vgg6_mnist} 
           \end{subfigure}
           
           \begin{subfigure}[b]{\textwidth}
                \centering
                \includegraphics[width=\textwidth]{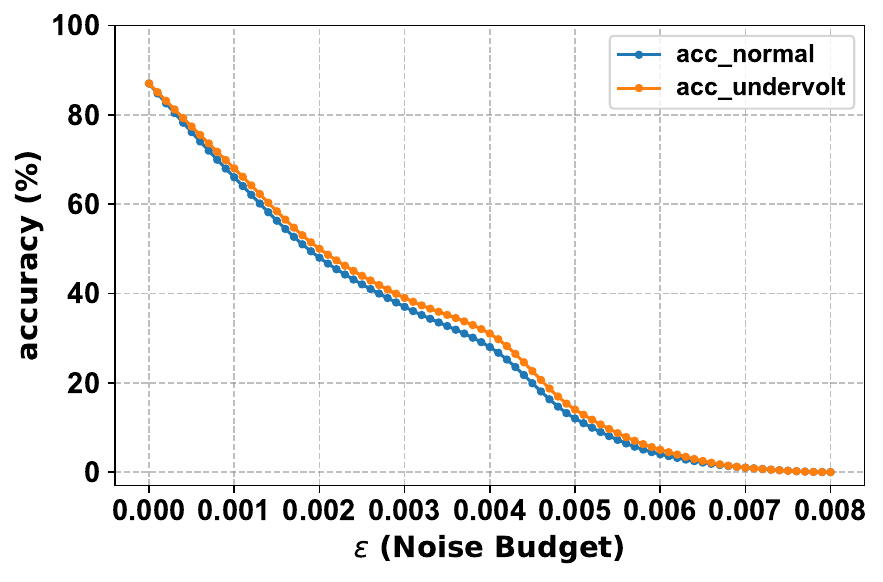}
                \caption{MobileNet-V3-CIFAR-10}
                \label{fig:at_bitflip_distribution_1_0} 
           \end{subfigure}}
           
    \caption{Comparison of the robustness of adversarial-trained models trained under normal GPU operation and undervolted GPU conditions using adversarial samples generated with various noise budgets}
    \label{fig:at_undervolting_training}
\end{figure*}

We then repeated the pipeline on the same datasets with the GPU held at the
lowest voltage at which the system remained stable, producing what we call the
\textit{undervolted model}.

Table~\ref{table:power_improvement} reports the voltage and frequency for
nominal and undervolted operation together with the resulting power
improvement. Frequency is unchanged between the two settings, so the reduction
is attributable to voltage alone.

\begin{table}[h!]
\centering
\caption{Voltage, Frequency, and Power Improvement Across Models}
\label{table:power_improvement}
\resizebox{\linewidth}{!}{
\begin{tblr}{
  colspec ={
    |Q[1.2cm, valign=m] 
    | Q[1.5cm,valign=m] 
    | Q[1.5cm,valign=m]
    | Q[1.5cm,valign=m]
    | Q[1.5cm,valign=m]|
  },
  cell{1-Z}{1-Z} = {c,m},
  hline{1,2}={1-Z}{solid}
}
Model & Nominal Voltage (V) & Undervolted Voltage (V) & Frequency (MHz) & Power Improvement (\%) 
\\\hline
LeNet   & 1.031 & 0.806 & 1980 & 38.8
\\\hline
VGG-6   & 1.031 & 0.837 & 1980 & 33.9
\\\hline
MobileNet & 1.031 & 0.806 & 1980 & 38.8
\\\hline

\end{tblr}}
\end{table}

Turning to robustness, Figure~\ref{fig:undervolting_training} shows that the gap between the two training regimes widens as the adversarial budget grows; baseline models lose accuracy rapidly under stronger perturbations, consistent with the documented vulnerability of CNNs, while undervolted models retain higher accuracy across all budgets at which either retains nonzero accuracy. The trend holds across every architecture and dataset evaluated, so controlled hardware-level noise introduced during training acts as an implicit regularizer that improves adversarial robustness, perhaps by favoring flatter minima or by exposing the network to fluctuations that resemble adversarial distortion. 

The gain was not uniform across different models, however, it was noticeably smaller for larger, deeper models such as MobileNetV3 (Figure~\ref{fig:bitflip_distribution_1_0}). We attribute this to greater fault masking in their deeper convolutional and intermediate computations, where the many nonlinear stages and pooling operations tend to absorb small perturbations before they propagate to the output. This interpretation is reinforced by our fault characterization, which shows that most observed faults occurred in lower-significance bits (Figure~\ref{fig:bitflip_distribution}), whose corruption is more easily tolerated by an already redundant representation. As future work, we plan to evaluate larger models with a higher proportion of linear layers, which offer fewer masking opportunities, to better understand how this robustness behavior generalizes across network depth and structure.

\subsection{Undervolted Adversarial Training}

To test the complementarity hypothesized in Section~\ref{sec:procedure}, we performed adversarial training under nominal and undervolted conditions. Both runs followed the standard PGD procedure: adversarial examples were generated on-the-fly from the model's current parameters at each iteration and used to update the weights, keeping the model exposed to perturbations throughout training. 

We then evaluated both models across a wide range of PGD attack strengths.
Figure~\ref{fig:at_undervolting_training} compares the two. The undervolted
version is consistently more robust across every architecture and dataset,
with the margin widening at medium to high perturbation budgets where the
baseline begins to deteriorate. The gain is notable given that adversarial
training already ranks among the strongest available defenses, yet
undervolting-induced noise improves it further.

\section{Discussion and future work}
The proposed undervolting-based approach demonstrated noticeable improvements over the baseline, highlighting its effectiveness as a promising technique for enhancing both the robustness and energy efficiency of AI models. Although the current results are preliminary, they provide strong evidence that undervolting can be leveraged not only as a power optimization technique but also as a mechanism for improving model resilience, introducing a new research direction at the intersection of energy-efficient computing and AI robustness.

Neural networks and AI now serve as the backbone of a wide range of application domains ~\cite{hariri2026test, hadi2025embedding}, so a hardware-level defense is only valuable insofar as it transfers beyond the architectures evaluated here. Building on these encouraging results, several directions will be explored in future work. First, we plan to extend the evaluation to larger and more complex foundation models particularly those with a higher proportion of linear layers, which offer fewer masking opportunities as well as quantized and other optimized AI models, to assess the generality and scalability of the approach. We also intend to investigate more advanced voltage and frequency scaling strategies to identify optimal operating points that maximize robustness while preserving computational efficiency. Another important direction is the characterization and localization of undervolting-induced bit-flips. By analyzing where bit-flips occur, their frequency, and their impact on model behavior, we aim to better understand fault propagation under undervolting conditions and support the development of more targeted and effective undervolting strategies.

\section{Conclusion}

This study demonstrates that controlled GPU undervolting offers a promising avenue for enhancing the robustness of CNNs while simultaneously reducing GPU energy consumption. By characterizing the stochastic faults induced by lowering voltage on real GPUs, we show that hardware-level perturbations can act as an implicit regularizer, improving generalization and resilience to both natural and adversarial input variations, without significant degradation in model accuracy. This work opens a new perspective at the intersection of GPU hardware and machine learning, motivating future research on larger foundation models.



\bibliographystyle{IEEEtran}
\bibliography{sample-base}

\end{document}